\documentclass[conference,a4paper]{IEEEtran}
\IEEEoverridecommandlockouts
\usepackage{cite}
\usepackage{amsmath,amssymb,amsfonts}
\usepackage{algorithmic}
\usepackage{graphicx}
\usepackage{textcomp}
\usepackage{xcolor}
\def\BibTeX{{\rm B\kern-.05em{\sc i\kern-.025em b}\kern-.08em
    T\kern-.1667em\lower.7ex\hbox{E}\kern-.125emX}}
\begin{document}


\title{Optimal Operating-Point Selection for Single-Linearization Trajectory Planning of Nonlinear Robot Manipulators
}

\author{\IEEEauthorblockN{Bhanuka Dayawansa}
\IEEEauthorblockA{\textit{Department of Electronic and Telecomm. Engineering} \\
\textit{University of Moratuwa}\\
Katubedda, Sri Lanka \\
dayawansasb.22@uom.lk}
\and
\IEEEauthorblockN{Rohan Munasinghe}
\IEEEauthorblockA{
\textit{Department of Electronic and Telecomm. Engineering} \\
\textit{University of Moratuwa} \\
Katubedda, Sri Lanka \\
\textit{Robotics Laboratory, Narva College} \\
\textit{University of Tartu} \\
Tartu 50090, Estonia \\
rohan@uom.lk
}

}

\maketitle

\begingroup
\renewcommand{\thefootnote}{}
\footnotetext{%
\copyright~2026 IEEE. Personal use of this material is permitted.
Permission from IEEE must be obtained for all other uses, in any current or future media, including reprinting/republishing this material for advertising or promotional purposes, creating new collective works, for resale or redistribution to servers or lists, or reuse of any copyrighted component of this work in other works.
}
\endgroup

\begin{abstract}
Single-point linearization provides a computationally efficient basis for optimization-based trajectory planning of nonlinear robot manipulators. However, planning performance depends on the selected linearization point. This paper presents an offline performance-driven approach that treats the location of a single linearization as an optimization variable. Candidate operating points are sampled from a common cubic seed trajectory, and each corresponding affine model is used to solve an identical constrained finite-horizon quadratic program (QP). The resulting control sequences are executed on the original nonlinear dynamics, and the operating point minimizing the nonlinear trajectory objective is selected through grid search and bounded refinement. Evaluation on four point-to-point motions of a three-degree-of-freedom UR5 manipulator yields optimal normalized operating-point locations between 0.325 and 0.935, demonstrating trajectory-dependent selection. Compared with the best-performing conventional initial, midpoint, or terminal linearization, the proposed approach reduces the nonlinear trajectory objective by 17–18\%, 21–22\%, 37–38\%, and 7–8\% across the four tasks, respectively. Additional nonlinear execution with identical PID feedback correction confirms that the performance advantage persists for a representative task. These results demonstrate that systematic operating-point selection can substantially improve single-linearization trajectory planning while preserving a fixed affine planning model.
\end{abstract}

\renewcommand\IEEEkeywordsname{Keywords}


\section{Introduction}

Robot manipulators are inherently nonlinear dynamical systems due to effects such as configuration-dependent inertia, Coriolis and centrifugal forces, and gravitational loading \cite{b1}. Consequently, trajectory planning and tracking control can be formulated either by explicitly accounting for the nonlinear system dynamics or by employing locally linearized models around selected operating points or nominal trajectories. Although nonlinear formulations can provide a more accurate representation of the manipulator dynamics, they often involve greater computational complexity\cite{b2,b3}. Furthermore, linear or affine manipulator models remain attractive because they enable optimization-based trajectory generation and control of manipulators such as linear quadratic regulation (LQR), Model Predictive control (MPC) and related methods \cite{b4}. Such formulations are especially useful when computational simplicity, predictable solver behavior, or repeated solution of constrained planning problems is preferred.

A common approach in manipulator trajectory planning is to linearize the nonlinear dynamics at a single operating point \cite{b5,b6,b7}. However, since the dynamics vary with joint configuration and velocity, different operating points yield different local models \cite{b8,b9}, raising the question of where along the motion linearization should occur.

Conventional choices include the initial or terminal state \cite{b10,b11}, where zero joint velocities simplify the dynamics, or the trajectory midpoint, as adopted in \cite{b7}. However, these heuristic choices may not adequately represent the nonlinear dynamics or account for how the selected model influences the optimization problem and its resulting trajectory and control sequence \cite{b10,b11}. Moreover, minimizing local linearization error does not necessarily yield the best finite-horizon nonlinear performance \cite{b12,b13}. Related energy-aware studies further show that comparable tracking performance does not imply equivalent actuator-energy requirements \cite{b14}, emphasizing the need to evaluate performance beyond trajectory accuracy.

This paper investigates the hypothesis that the operating point used for linearization should itself be treated as a design variable. The proposed framework parameterizes the location of a single operating point by a normalized variable
\[
\lambda \in [0,1],
\]
where the operating point is sampled from a common seed trajectory at time
\[
\bar{t} = \lambda T_f.
\]
For each candidate value of $\lambda$, exactly one affine linear model is constructed and the same constrained finite-horizon quadratic program is solved. The resulting control sequence is then applied to the original nonlinear robot model, and the corresponding nonlinear trajectory cost is used as an outer performance measure. The operating point is selected offline as the value of $\lambda$ that minimizes this nonlinear open-loop trajectory cost.

Thus, the proposed approach does not replace the single-linearization planner with a fully nonlinear or successively linearized optimization method. Instead, it preserves the simplicity and computational efficiency of a single affine planning model while eliminating the arbitrary selection of its operating point. The planner implementation explicitly uses an outer nonlinear-performance optimization around a common inner QP, while $\lambda = 0$, $0.5$, and $1$ are retained only as initial-, midpoint-, and terminal-linearization benchmarks respectively.

The main contributions of this work are threefold. First, the operating point location of a single robot linearization is formulated as an explicit optimization variable rather than a predetermined modeling choice. Second, we empirically demonstrate that the best operating point is strongly trajectory dependent, and therefore no static, single initial, midpoint, or terminal-based choice is consistently optimal. Third, the selected optimal operating point is validated by executing the generated trajectories on the nonlinear robot model both in open loop and with common feedback correction using a PID tracking controller, showing that the effect extends beyond the predicted behavior of the linear planning model. Existing approaches improve model fidelity by repeatedly linearizing along a trajectory, whereas this work tries to answer a different question: if computational constraints require a single fixed linearization, where should that linearization be performed?

\section{Problem Formulation}

\subsection{Nonlinear Robot Dynamics}

The Manipulator Dynamic Model can be represented by
\begin{equation}
D(q)\ddot{q}+C(q,\dot{q})\dot{q}+G(q)=\tau_{\mathrm{abs}},
\label{eq:manipulator_dynamics}
\end{equation}
where $q$, $\dot{q}$, and $\ddot{q}$ denote joint position, velocity, and acceleration, respectively, $D(q)$ is the inertia matrix; $C(q,\dot{q})\dot{q}$ is the Coriolis and centrifugal effects term which contain velocity-dependent dynamic terms. $G(q)$ represents the gravity vector and $\tau_{\mathrm{abs}}$ is the absolute actuator torque.

The gravity compensated torque used in this study could be defined as,
\begin{equation}
\tau_{\mathrm{abs}}=G(q)+u,
\label{eq:gravity_comp_torque}
\end{equation}
which gives
\begin{equation}
D(q)\ddot{q}=u-C(q,\dot{q})\dot{q}.
\label{eq:gravity_comp_dynamics}
\end{equation}

This input convention is used consistently in both the planning and nonlinear execution stages.

The state vector is defined relative to the desired terminal configuration as
\begin{equation}
x=
\begin{bmatrix}
q-q_g\\
\dot{q}-\dot{q}_g
\end{bmatrix},
\label{eq:state_vector}
\end{equation}
where $q$ denotes the current joint configuration and $q_g$ denotes the goal configuration.

The resulting nonlinear state equation can be obtained as,
\begin{equation}
\dot{x}=f(x,u).
\label{eq:nonlinear_state}
\end{equation}

\subsection{Parameterization of the Linearization Point}

A common cubic trajectory is first defined between the start and goal conditions as a seed trajectory. Rather than changing the seed trajectory for different candidate models, all candidate linearizations are sampled from this same trajectory.

The operating time is parameterized as
\begin{equation}
\bar{t}(\lambda)=\lambda T_f,\quad 0\leq\lambda\leq1.
\label{eq:operating_time}
\end{equation}

The corresponding state and input are denoted as
\begin{equation*}
\bar{x}(\lambda),\ \bar{u}(\lambda).
\end{equation*}

The operating-point position, velocity, and acceleration are obtained by sampling the same endpoint-constrained cubic seed trajectory at $\bar{t}=\lambda T_f$:
\begin{equation*}
\bar{q}=q_s(\bar{t}),\quad
\dot{\bar{q}}=\dot{q}_s(\bar{t}),\quad
\ddot{\bar{q}}=\ddot{q}_s(\bar{t}),
\end{equation*}
where the cubic trajectory satisfies the prescribed initial and terminal position and velocity conditions. Consistent with the gravity-compensated input convention in (2)--(3), the corresponding operating-point input is reconstructed from the nonlinear manipulator dynamics as
\begin{equation}
\bar{u}(\lambda)
=
D(\bar{q})\ddot{\bar{q}}
+
C(\bar{q},\dot{\bar{q}})\dot{\bar{q}}.
\label{eq:operating_input}
\end{equation}

Thus, both the state and input used for linearization are obtained consistently from the same cubic seed trajectory.

Linearization of the nonlinear state equation (5) around this point gives
\begin{equation}
\dot{x}
\approx
A_c(\lambda)x
+
B_c(\lambda)u
+
c_c(\lambda),
\label{eq:linearized_continuous}
\end{equation}
Where $A_c$ and $B_c$ can be obtained using the Taylor-Jacobian Linearization
\begin{equation}
A_c
=
\left.
\frac{\partial f}{\partial x}
\right|_{\bar{x},\bar{u}},
\quad
B_c
=
\left.
\frac{\partial f}{\partial u}
\right|_{\bar{x},\bar{u}},
\label{eq:jacobians}
\end{equation}
and the affine term $c_c$ ensures consistency at the operating point.

The implementation samples all principal operating points from the same cubic seed trajectory, which avoids changing the reference used to define the candidate linearizations.

After exact zero-order-hold discretization the discretized model can be depicted as,
\begin{equation}
x_{k+1}
=
A(\lambda)x_k
+
B(\lambda)u_k
+
g(\lambda).
\label{eq:discrete_affine}
\end{equation}

Only one such model is used throughout the prediction horizon for each candidate $\lambda$.

\section{Performance-Driven Operating-Point Selection}

\subsection{Inner Trajectory-Planning Problem}

For a fixed $\lambda$, the affine system in (10) is used to construct a finite-horizon quadratic program.

The objective function has the general form
\begin{equation}
J_{\mathrm{QP}}
=
\sum_{k=0}^{N-1}
\left(
x_k^{\top}Q_xx_k
+
u_k^{\top}R_uu_k
\right),
\label{eq:qp_objective}
\end{equation}
subject to the affine system dynamics, initial conditions, actuator limits, and terminal goal constraints. Here $Q_x$ is the state-cost matrix and $R_u$ is the control cost matrix, The same horizon, objective weights, constraints, and solver settings are used for every candidate operating point to maintain fairness. Consequently, the parameter intentionally varied between candidates is the location of the single linearization.

The QP solution corresponding to a particular candidate is denoted by
\begin{equation}
U^*(\lambda)
=
\left[
u_0^*
\ \cdots\
u_{N-1}^*
\right].
\label{eq:qp_solution}
\end{equation}

\subsection{Outer Nonlinear Performance Evaluation}

A control sequence that performs well under the affine planning model does not necessarily produce the same behavior when executed on the original nonlinear dynamics of the manipulator. Hence, each optimized sequence $U^*(\lambda)$ is applied to the nonlinear robot model and evaluated.

The nonlinear rollout is evaluated using the same quadratic state and control weighting used in the planning problem. Let
\begin{equation*}
x_{\mathrm{NL},k}
=
\begin{bmatrix}
q_{\mathrm{NL},k}-q_g\\
\dot{q}_{\mathrm{NL},k}-\dot{q}_g
\end{bmatrix}.
\end{equation*}

The nonlinear open-loop task objective is defined as
\begin{equation}
\begin{aligned}
J_{\mathrm{NL}}(\lambda)
&=
\frac{1}{2}
\sum_{k=0}^{N}
x_{\mathrm{NL},k}^{\top}
Q_x
x_{\mathrm{NL},k}
\\
&\quad
+
\frac{1}{2}
\sum_{k=0}^{N-1}
u_k^*(\lambda)^{\top}
R_u
u_k^*(\lambda),
\end{aligned}
\label{eq:nonlinear_objective}
\end{equation}
where $Q_x=\mathrm{blkdiag}(Q_q,Q_{\dot{q}})$, with $Q_q=150I_3$, $Q_{\dot{q}}=10I_3$, and $R_u=4I_3$. The nonlinear states are obtained by applying the QP-generated input $u_k^*(\lambda)$to the original nonlinear dynamics using a fourth-order Runge--Kutta integration with sampling interval $\Delta t=0.02\ \mathrm{s}$, with the input held constant over each sampling interval.

The operating-point-selection problem becomes
\begin{equation*}
\lambda^*
=
\underset{\lambda\in[0,1]}{\arg\min}
J_{\mathrm{NL}}(\lambda)
\end{equation*}
which forms the central optimization problem investigated in this work.

Accordingly, the proposed method is organized into two nested optimization levels i.e. an inner constrained trajectory planning problem and an outer scalar operating-point selection problem.

\subsection{Offline Search Strategy}

The operating point parameter $\lambda$ is initially evaluated over the interval $0\leq\lambda\leq1$ using a coarse global grid,
\begin{equation}
\lambda
\in
\{0\ ,\ 0.05\ ,\ 0.10\ldots1.0\}.
\label{eq:lambda_grid}
\end{equation}

The minimum-cost grid point is further refined using bounded derivative-free optimization. A derivative-free search is appropriate since the outer objective contains the solution of a constrained QP, and changes in the active constraint set can make the cost nonsmooth with respect to $\lambda$. The implemented search uses a 0.05 global grid which is then followed by \texttt{fminbnd} refinement.

Three conventional operating points which are commonly used for linearization are retained for comparison and are treated strictly as heuristic baselines:
\begin{equation}
\lambda_{\mathrm{initial}}=0,\quad
\lambda_{\mathrm{mid}}=0.5,\quad
\lambda_{\mathrm{terminal}}=1.
\label{eq:heuristic_points}
\end{equation}

\section{Experimental Setup}

The study is conducted using a three-degree-of-freedom model of the Universal Robots UR5 manipulator. Joints 1--3 are actively modeled while joints 4--6 remain fixed. The simulation duration is taken as
\begin{equation*}
T_f=2.0\ \mathrm{s},
\end{equation*}
with sampling interval
\begin{equation*}
\Delta t=0.02\ \mathrm{s},
\end{equation*}
Which gives $N=100$ control intervals.

The state cost weights (joint-position and velocity) are set as
\begin{equation*}
Q_q=150I_3,\quad
Q_{\dot{q}}=10I_3,
\end{equation*}
such that
\begin{equation*}
Q_x
=
\begin{bmatrix}
Q_q & 0\\
0 & Q_{\dot{q}}
\end{bmatrix}.
\end{equation*}
while the control cost weighting is
\begin{equation*}
R_u=4I_3.
\end{equation*}

Gravity-compensated actuator inputs are constrained to
\begin{equation*}
-60\ \mathrm{Nm}
\leq
u_i
\leq
60\ \mathrm{Nm}.
\end{equation*}

These settings correspond to the optimal trajectory planner configuration used to generate the supplied results.

Four point-to-point trajectories are selected for this study which are denoted as T1--T4, and are used to evaluate whether a common preferred linearization point exists across different motions.

\begin{table}[t]
\caption{Start and Goal Joint Configurations for the Evaluated
Point-to-Point Trajectories}
\label{tab:trajectory_tasks}
\centering
\footnotesize
\setlength{\tabcolsep}{3pt}
\renewcommand{\arraystretch}{1.10}
\begin{tabular}{c c c c}
\hline
\textbf{Task} & $\mathbf{q}_0$ [rad] & $\mathbf{q}_g$ [rad] & $T_f$ [s] \\
\hline
T1 & $[-1.00,\,-1.40,\,2.30]$ & $[1.00,\,-0.25,\,0.80]$ & 2.0 \\
T2 & $[-0.60,\,-1.20,\,2.10]$ & $[0.60,\,-0.45,\,1.20]$ & 2.0 \\
T3 & $[-0.90,\,-1.45,\,0.90]$ & $[0.90,\,-0.35,\,2.20]$ & 2.0 \\
T4 & $[-0.40,\,-1.30,\,2.50]$ & $[0.40,\,-1.00,\,0.70]$ & 2.0 \\
\hline
\end{tabular}
\end{table}

For the nonlinear execution of the planned trajectory, identical PID gain parameters are used for all operating points considered.


\section{Results and Discussion}
\label{sec:results}

\subsection{Sensitivity of Trajectory Quality to the Linearization Point}

\begin{figure}[t]
    \centering
    \includegraphics[width=\columnwidth]{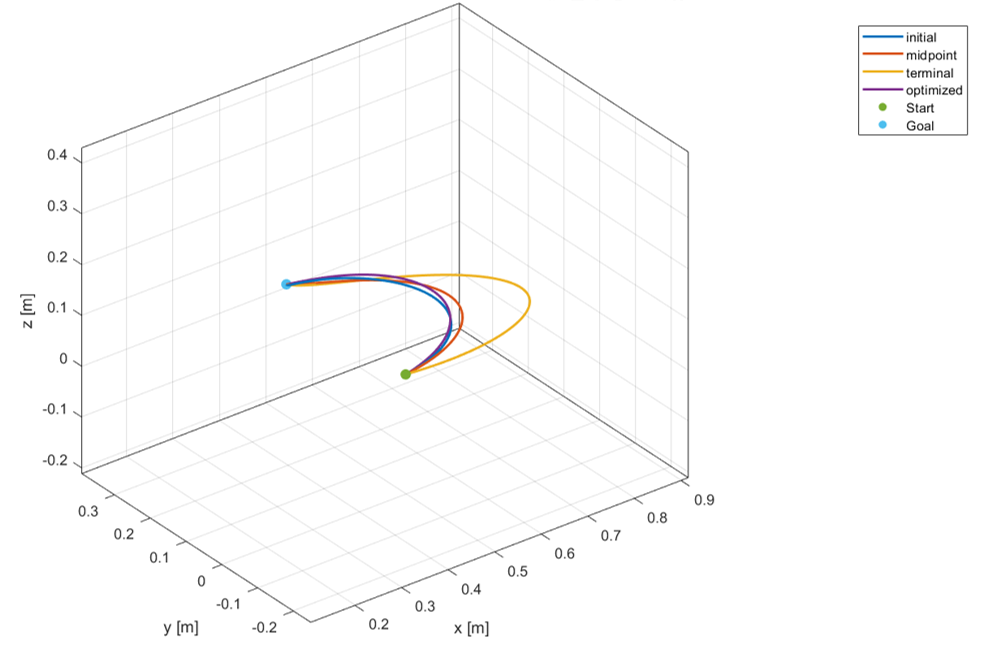}
    \caption{Three-dimensional end-effector trajectories obtained using different single-linearization operating points.}
    \label{fig:ee_paths}
\end{figure}

The first important observation is that the nonlinear trajectory quality varies
considerably based on the operating point location. This could be observed when
the planned point-to-point trajectories are executed in the nonlinear UR5
manipulator model under identical conditions. This could be observed in the
visual representation shown in Fig.~\ref{fig:ee_paths} of the task-space
trajectories the end-effector of the UR5 manipulator takes when tracking the
planned joint-space trajectory.

The only difference which causes this trajectory deviation is varying the
operating point used for linearization of the robot model. As it has been
mentioned in the experimental setup, an identical objective function, cost
weights, tracking gain parameters, and start and goal joint configurations
have been used.

\subsection{The Optimal Linearization Point Is Trajectory Dependent}

\begin{figure}[t]
    \centering
    \includegraphics[width=\columnwidth]{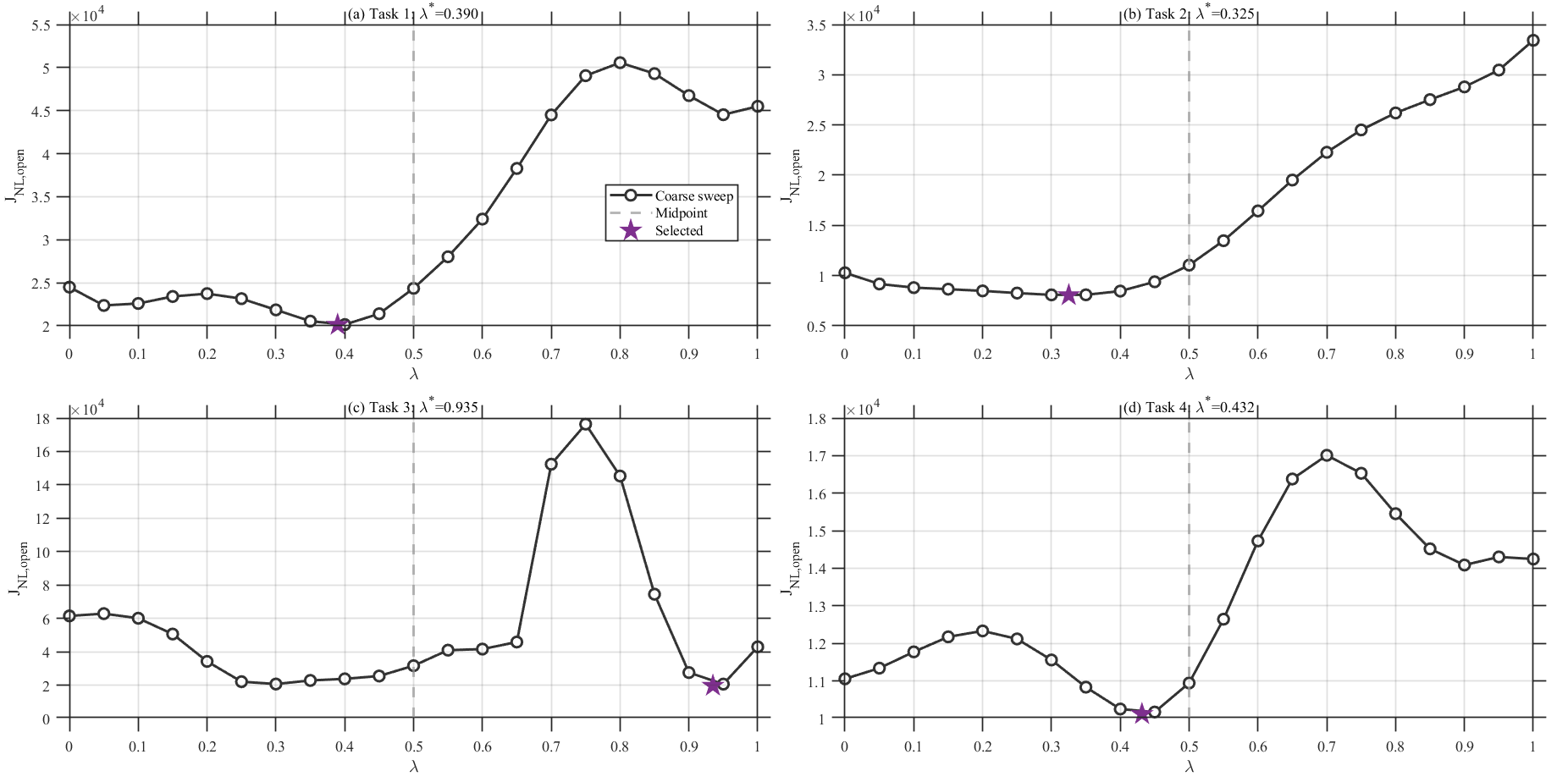}
    \caption{Nonlinear open-loop trajectory cost as a function of the
    normalized single-linearization location $\lambda$ for four robot motions.}
    \label{fig:cost_lambda}
\end{figure}

Fig.~\ref{fig:cost_lambda} shows how the nonlinear trajectory quality and the
optimal operating point location varies when the trajectory changes. The
conventional midpoint operating point, i.e., when $\lambda = 0.5$, is indicated
for reference while the initial and terminal operating points are at
$\lambda = 0$ and $\lambda = 1$, respectively. The star sign denotes
$\lambda^{\ast}$, which is the performance-selected optimal operating point.

As it could be observed, the optimal operating point location varies
significantly between varying trajectories which are labelled as tasks. The
four experiments give the optimal operating points as
\begin{equation}
\lambda_{T1}^{\ast} = 0.390,\quad
\lambda_{T2}^{\ast} = 0.325,\quad
\lambda_{T3}^{\ast} = 0.935,\quad
\lambda_{T4}^{\ast} = 0.432.
\end{equation}

This demonstrates trajectory-dependent linearization-point selection and
provides evidence against adopting a universally fixed operating point for
linearization. The trajectories $T1$, $T2$, and $T4$ favour operating points
obtained from within the first 30\%--45\% of the motion while $T3$ prefers its
optimum near the end of the trajectory at $\lambda^{\ast} = 0.935$.

More importantly, it could also be observed that the cost landscapes differ
qualitatively. For example, $T2$ demonstrates a relatively smooth increase
after its minimum at $\lambda = 0.325$, whereas $T3$ demonstrates a highly
nonuniform cost landscape and eventually favours a minimum close to the
terminal part of the motion at $\lambda^{\ast} = 0.935$.

Hence, it indicates that despite the robot model, planning formulation,
horizon, and the objective function used for optimization being unchanged,
changing the motion task can move the optimal operating point over a
substantial portion of the trajectory. This observation supports the central
hypothesis of this paper, i.e., the appropriate single linearization point of a
nonlinear robotic system is a property of the trajectory planning problem
rather than a universal robot modeling constant.

\subsection{Performance Relative to Conventional Linearization Heuristics}

The practical relevance of optimizing $\lambda$ depends not only on whether
the optimal operating point differs from a universally defined constant
operating point such as the initial, terminal, or the midpoint, but also on how
much performance degradation takes place by choosing such conventional
operating points.

To quantify this performance degradation, we introduce a parameter $R$, which
is the regret relative to the selected point $\lambda^{\ast}$:

\begin{equation}
R(\lambda)
=
100
\frac{
J_{\mathrm{NL}}(\lambda)
-
J_{\mathrm{NL}}(\lambda^{\ast})
}{
J_{\mathrm{NL}}(\lambda^{\ast})
}.
\label{eq:regret}
\end{equation}

\begin{figure}[t]
    \centering
    \includegraphics[width=\columnwidth]{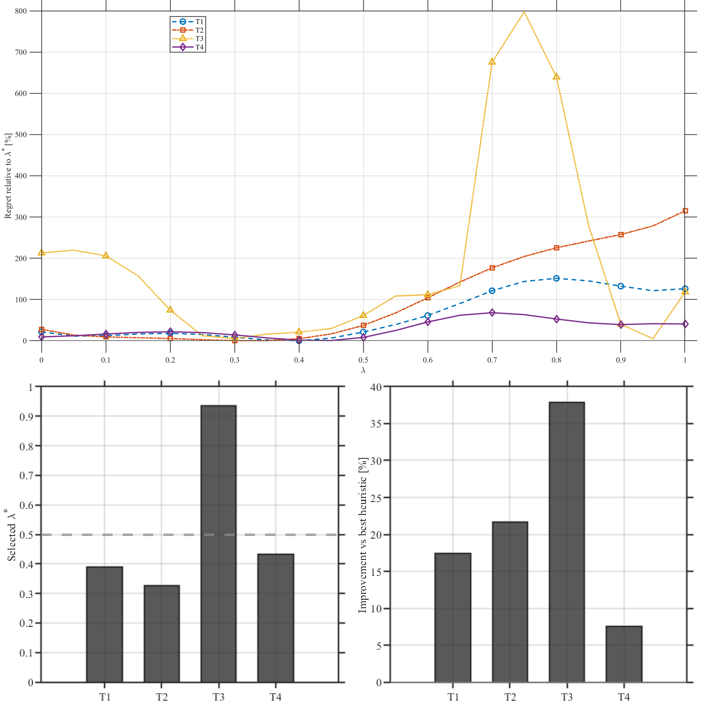}
    \caption{Summary of operating-point optimization across four motion tasks.}
    \label{fig:optimization_summary}
\end{figure}

The top plot in Fig.~\ref{fig:optimization_summary} shows that the regret
$R(\lambda)$ can vary greatly between tasks and can become very large during
parts of the motion where the nonlinearity becomes strongly prominent. The
left panel highlights wide variation of $\lambda^{\ast}$ across trajectories,
while the right panel goes further by comparing the performance of the
computed optimal operating point against the best performing operating point
of the three conventional universal operating points, rather than against a
deliberately poor baseline.

From the plotted results, it could be observed that the performance reduction
relative to the best conventional heuristic is approximately
\begin{align*}
17\%-18\% &\quad \text{for } T1,\\
21\%-22\% &\quad \text{for } T2,\\
37\%-38\% &\quad \text{for } T3,\\
7\%-8\%   &\quad \text{for } T4.
\end{align*}

Consequently, all four motion tasks benefit from explicit optimal
operating-point selection, although the magnitude of the benefit is trajectory
dependent. This comparison is particularly significant as the proposed method
is not merely outperforming an arbitrary baseline, but for each task the
baseline is selected as whichever of the initial, midpoint, or terminal
linearizations already performs best.

The remaining performance improvement therefore quantifies the benefit
obtained by treating $\lambda$ as a continuous design variable.

\subsection{Cumulative Cost and Cost Decomposition}

\begin{figure}[t]
    \centering
    \includegraphics[width=\columnwidth]{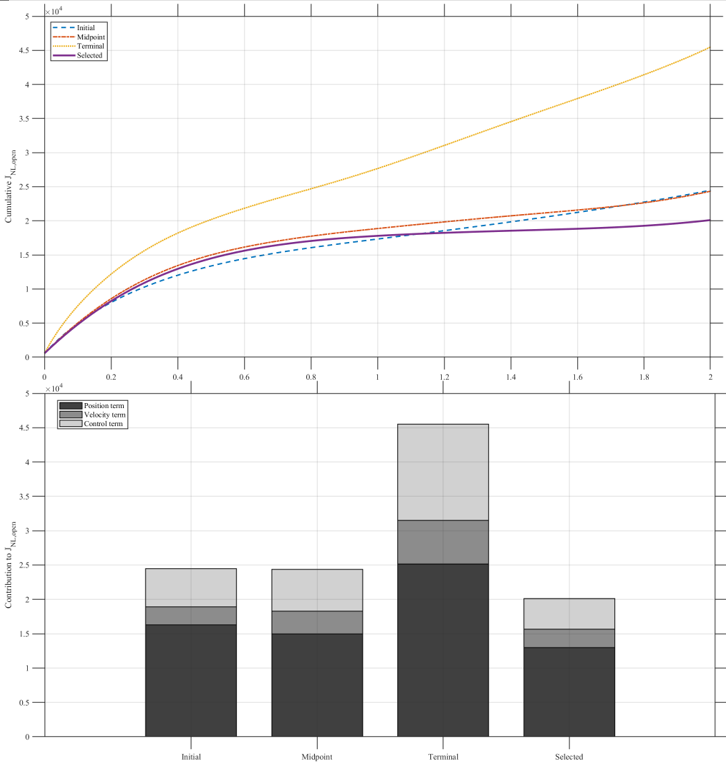}
    \caption{Cumulative cost and cost decomposition of the nonlinear trajectory
    objective for the initial, midpoint, terminal, and optimal operating
    points.}
    \label{fig:cumulative_cost}
\end{figure}

Fig.~\ref{fig:cumulative_cost} indicates that the cumulative cost based on the
nonlinear trajectory objective when the optimal operating point is selected
for the linearization is comparatively the lowest compared to the cases where
the conventional operating points are selected for linearization. This
indicates that the benefit develops over the finite motion rather than
appearing only at the final sample.

Further, the decomposition of the nonlinear trajectory objective into
position, velocity, and control contributions for varying operating points
has been demonstrated. It can be observed that the optimal operating point
produces the lowest overall objective for the representative task.
Interestingly, the lowest control cost is reported for the optimal
linearization, which suggests that it has a positive effect on reducing
control effort.

\subsection{Nonlinear Execution With Feedback Correction}

To validate the planner stage results, the planned trajectories were executed
in a nonlinear UR5 robot manipulator model. Before evaluating closed-loop
correction, it was verified that the execution implementation reproduces the
planner's nonlinear rollout. An independent planner-executor reproduction
check produced relative cost differences on the order of $10^{-13}$,
confirming numerical consistency of the nonlinear rollout implementation.

However, the open-loop nonlinear objective is the criterion used to select the
optimal operating point $\lambda^{\ast}$. Therefore, lower open-loop cost for
the selected candidate is an expected outcome of the optimization procedure.
A stronger validation would be to identify if the optimal linearization
benefit persists when the planned trajectory input is embedded in a common
feedback architecture.

Hence, the execution stage evaluates two relevant cases:

\begin{equation}
u = u_{\mathrm{plan}}
\label{eq:open_loop_execution}
\end{equation}

and

\begin{equation}
u = u_{\mathrm{plan}} + u_{\mathrm{PID}}.
\label{eq:pid_execution}
\end{equation}

The same PID gains and actuator limits are used for every operating point and
the execution framework deliberately separates open-loop planning quality from
planner-feedforward-plus-feedback behavior.

\begin{figure}[t]
    \centering
    \includegraphics[width=\columnwidth]{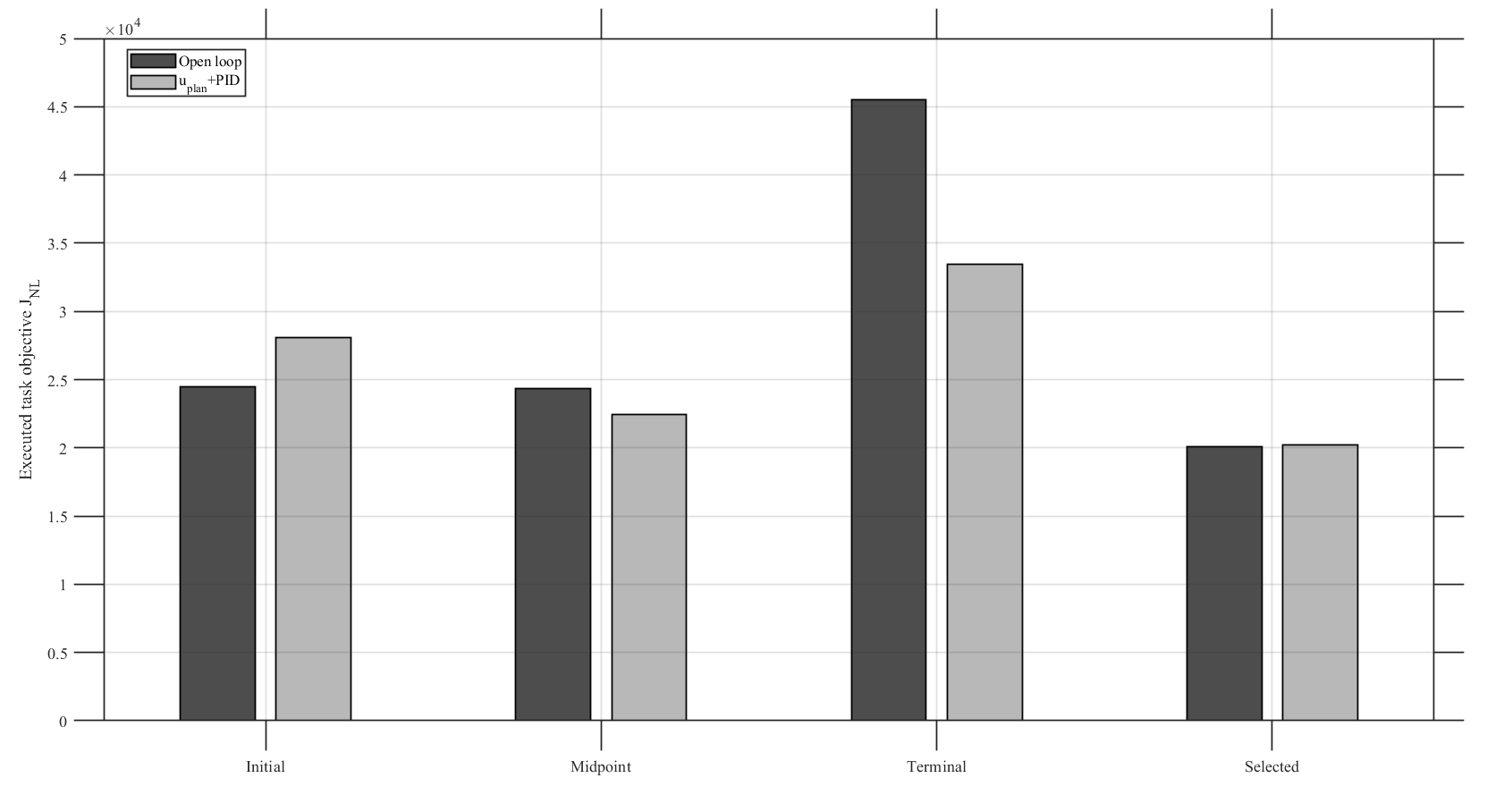}
    \caption{Nonlinear executed task objective for initial, midpoint, terminal,
    and selected operating points under open-loop application of the planned
    input and planner-feedforward-plus-PID execution.}
    \label{fig:nonlinear_execution}
\end{figure}

Fig.~\ref{fig:nonlinear_execution} indicates that the computed optimal
operating point gives the lowest executed objective in both cases for the
representative validation trajectory selected. Most importantly, the selected
operating point remains favourable after the PID feedback controller is
introduced.

This indicates that the improvement due to the linearization-point
optimization is not limited to the numerical prediction observed for the
linear planning model. This shows that although nonlinear open-loop cost was
explicitly used to compute $\lambda^{\ast}$, the selected plan retains its
advantage when combined with identical PID feedback control, providing an
additional validation of the operating point choice.

\section{Conclusion}
This paper presented a performance-driven method for selecting the operating point of a single affine linearization used in nonlinear manipulator trajectory planning. By minimizing the nonlinear execution cost of trajectories generated through constrained quadratic programming, the proposed method eliminates the need for a predetermined linearization location. Evaluation on four UR5 motions revealed strongly trajectory-dependent optimal locations, ranging from 0.325 to 0.935, with objective reductions of approximately 7–38\% relative to the best conventional initial, midpoint, or terminal linearization. Additional nonlinear execution with identical PID feedback correction preserved the performance advantage for a representative task. These findings demonstrate that optimizing the linearization location can improve trajectory planning while retaining a fixed affine model throughout the planning horizon.

\section*{Acknowledgment}
The co-author was partially supported by the project
“Increasing the Knowledge Intensity of Ida-Viru Entrepreneurship,”
co-funded by the European Union.

\end{document}